# Predicting Aircraft Trajectories: A Deep Generative Convolutional Recurrent Neural Networks Approach

Yulin Liu

Institutes of Transportation Studies, University of California, Berkeley, liuyulin101@berkeley.edu

Mark Hansen

Institutes of Transportation Studies, University of California, Berkeley, mhansen@ce.berkeley.edu

**Abstract**

Reliable 4D aircraft trajectory prediction, whether in a real-time setting or for analysis of counterfactuals, is important to the efficiency of the aviation system. Toward this end, we first propose a highly generalizable efficient tree-based matching algorithm to construct image-like feature maps from high-fidelity meteorological datasets – wind, temperature and convective weather. We then model the track points on trajectories as conditional Gaussian mixtures with parameters to be learned from our proposed deep generative model, which is an end-to-end convolutional recurrent neural network that consists of a long short-term memory (LSTM) encoder network and a mixture density LSTM decoder network. The encoder network embeds last-filed flight plan information into fixed-size hidden state variables and feeds the decoder network, which further learns the spatiotemporal correlations from the historical flight tracks and outputs the parameters of Gaussian mixtures. Convolutional layers are integrated into the pipeline to learn representations from the high-dimension weather features. During the inference process, beam search, adaptive Kalman filter, and Rauch-Tung-Striebel smoother algorithms are used to prune the variance of generated trajectories.

## 1. Introduction

With the growing demand for air traffic, it is crucial to monitor and control air traffic flow to ensure the safety and efficiency of the National Airspace System (NAS). The Federal Aviation Administration (FAA) developed the Traffic Flow Management System (TFMS) to estimate sector traffic loads, make planning decisions, and evaluate historical performance. In Europe, EUROCONTROL adopted the PREDICT system ([1]) to forecast the pre-tactical traffic load. These systems depend on aircraft trajectory prediction tools. However, due to the computational complexity, most of today's prediction systems are based on deterministic flight trajectory prediction processes, using, for example, filed flight plans or historical routes ([2], [3]). However, demand forecasting based upon deterministic processes barely considers uncertainties such as unsteady weather conditions, which could lead to overestimating the traffic load for sectors with bad weather and induce unnecessary Traffic Management Initiatives (TMIs). From a system point of view, sector demand is not only the result of planned flight routes, but also reroutes in response to TMIs and weather changes. Furthermore, in the next generation of the NAS system (NextGen), the concept of trajectory based operations (TBO) has been proposed to be a cornerstone ([4]). TBO requires the development of tools to plan and predict accurate aircraft trajectories in four dimensions – lateral, vertical and time. In recent years, efforts have been spent on developing systems focusing on optimizing and controlling aircraft 4D trajectories ([5], [6], [7]), however, there are far fewer studies on predicting 4D trajectories. Therefore, in this paper, we propose a methodological framework to predict actual 4D trajectories, which incorporates different sources of uncertainties including weather, wind, and management actions.

In our framework, the future trajectory of an aircraft is modeled as a sequence of 4D coordinates that are correlated with its realized trajectory, last filed flight plan, which is a sequence of 2D waypoints, and weather conditions in the vicinity. Therefore, we can formulate our task as a "sequence to sequence learning" problem, in which the input sequence is the flight plan and the output is the actual flight trajectory. To solve this sequential learning problem, an encoder-decoder recurrent neural network structure has been employed, where the encoder learns from the flight plan and the decoder integrates the weather information and recursively "translates" the embedded flight plan information into a full 4D trajectory. We further assume the 4D trajectory coordinates follow conditional Gaussian mixtures, whose parameters are learned by the decoder. To model the weather effect, we first propose an efficient tree-based matching algorithm that can work both in *batch mode* and *recursive mode* to 4D match aircraft coordinates with nearby convective weather, wind speeds and air temperature. Then we integrate convolutional layers into the decoder network pipeline to extract representations from the high-dimension weather features. In summary, our paper has the following contributions.

- To the authors' knowledge, this paper is the first paper using an encoder-decoder recurrent neural network structure to predict 4D aircraft trajectories. Once trained, our model can also be easily adapted into systems that require real-time predictions with accurate prediction intervals.
- Our proposed approach is a generative model that incorporates multiple factors that influence trajectories, including convection, wind, and temperature. These features are constructed by an efficient tree-based matching algorithm, which is highly generalizable and can be used in many spatiotemporal matching applications.
- We propose both training and inference pipelines, in which the inference pipeline employs multiple filtering and searching algorithms to produce the best predictions.

The rest of the paper is organized as follows. In section 2, we review related works in trajectory prediction and deep neural networks. Section 3 introduces related concepts and summarizes our data sources and preprocessing procedure. Section 4 presents our matching (feature engineering) algorithms, and section 5 describes our methodological framework and model architecture. In section 6, we present the results of a case study. Section 7 offers conclusions and suggestions for future research.

## 2. Related work

Common approaches to aircraft trajectory prediction can be summarized into two categories: deterministic and probabilistic. The former approach ([8], [9], [10]) usually applies a specific aerodynamic model to estimate the state of an aircraft and then propagate the estimated states into the future (e.g., using Kalman filter). Although it accounts for specific aircraft parameters and kinematic equations, this approach, without considering any uncertainties such as future winds and pilot control actions, can either predict only a specific phase of a flight or suffers from degraded prediction accuracy. Moreover, this approach yields a "point estimate" of the future trajectory rather than a prediction internal. In contrast, the probabilistic approach ([11], [12], [13]) understates the aerodynamics but relies on statistical models to learn how aircraft fly from one point to another from historical trajectory datasets. Leege et al. ([11]) trained a generalized linear model (GLM) to use wind and aircraft initial state to predict aircraft trajectories in the arrival airport's terminal area. However, instead of predicting a 4D trajectory, the authors only predict the time of arrival to a fixed set of reference points (a.k.a., waypoints). Choi and Hebert ([12]) proposed a Markov model to predict future motion of a moving object based on its past movement. In their work, the trajectory of an object is broken down into a sequence of short segments, which are assumed to be generated from latent states. Therefore, their method predicts future trajectory segments of an object instead of actual coordinates. Ayhan and Samet ([13]) extended the work of Choi and Hebert and applied the method to predict actual 4D coordinates (track points) of an aircraft trajectory. In the paper, the 4D coordinate of the aircraft is defined as the hidden state, and the observed weather information that is closest to the aircraft coordinate is a realization of the hidden state. By training a hidden Markov model (HMM) on a historical trajectory and weather dataset, the authors obtained the parameters of the HMM model, which were further used to predict trajectories given the observed weather sequence. However, their model has three main drawbacks. First of all, the hidden state is fixed, indicating that every predicted track point can only be one of the historical track points. Second, for each track point, the authors only consider the weather condition that is closest to it, however, in aircraft routing problem, pilots usually consider a much larger region. Lastly, the prediction is highly dependent on the quality of the clustering results of the observed weather sequence, which is in practice difficult to control.

To a large extent, aircraft trajectories have clear temporal and spatial patterns, demonstrating high predictability. Thus, with the right tools, these patterns can be employed in predictive models. Recurrent neural networks (RNNs) have become the state of the art for sequence modeling. Furthermore, long short-term memory (LSTM) ([14]) is one of the most popular variations of RNNs with proven ability and stability in solving tasks in various domains, such as speech recognition ([15], [16], [17]), neural translation ([18]), and image captioning ([19], [20], [21]). The most relevant work is that of work of Lin et al. ([22]) and Alahi et al. ([23]). Lin et al. ([22]) proposed a deep generative model, which integrated an input-output HMM (IO-HMM) and a LSTM network, to model urban mobility. In the model, the daily activities of travelers are sequences of locations they visit through a day and are generated by the IO-HMM from cellular data. Then spatiotemporal patterns of the activities are learned by a LSTM network. Alahi et al. ([23]) proposed a "Social LSTM" model to predict human trajectories in crowded spaces. In the space, multiple individuals are observed, and each individual's trajectory is modeled by a LSTM network. To share information among individuals, the authors use the pooling techniques to connect those LSTM networks. Comparing to simple models such as Gaussian process and linear model, the authors approach has far better prediction errors on two open source datasets.

Inspired by Sutskever et al. ([18]), Lin et al. ([22]) and Alahi et al. ([23]), we develop an encoder-decoder LSTM-based generative model to predict aircraft 4D trajectories. Our approach differs from the past efforts in the following aspects: (a) it is a generative model that both learns the spatiotemporal trajectory patterns and can generate a full 4D trajectory given past information; (b) it uses high-dimension regional features generated from weather variables; (c) it gives the possibility of predicting different routes that are not observed in the training set.

## 3. Preliminaries

### 3.1. Definitions and Notations

*Definition 1.* A *4D trajectory* $TP = [p_1, p_2, ..., p_T]$, $p_t = [x, y, z, t]$ is defined as a sequence of 3-dimensional positions (longitude $x$, latitude $y$, altitude $z$) and time $t$.

*Definition 2.* A *pre-departure last filed flight plan* $\tilde{X} = [\tilde{X}_1, \tilde{X}_2, ..., \tilde{X}_{\tilde{T}}]$, $\tilde{X}_i = [x, y]$ is a sequence of 2-dimensional positions (longitude $x$, latitude $y$).

*Definition 3.* The *aircraft state* $X_t = [x, y, z, \dot{x}, \dot{y}]$ at time $t$ is defined as a tuple of aircraft 3D positions (longitude $x$, latitude $y$, altitude $z$) and lateral speeds (longitude speed $\dot{x}$, latitude speed $\dot{y}$).

*Definition 4. Georeferencing system* $G$ is a grid of fixed spatial points in the 3D space.

*Definition 5.* A *4D matching algorithm* is an algorithm to match a 4D trajectory point $p_t$ with datasets of interest with closest 4D distance (lateral, vertical and temporal).

*Definition 6.* A *feature cube* $F_t$ is a multi-dimensional array of 4D matched features around a 4D trajectory point $p_t$.

### 3.2. Problem Formulation

In this work, we seek to learn a generative model that predicts actual 4D aircraft trajectories. For each flight, we observe its last filed flight plan prior to departure $\tilde{X}$, which is a sequence of 2D flight tracks that guides the actual trajectory but is not always strictly followed. Given the aircraft state $X_t$ at *t*, we observe and 4D match the weather conditions such as wind speeds, air temperature and convective weather in the vicinity of the aircraft location, and we predict its state for the next time instance *t+1*. Thus, given flight plan, weather datasets, and some initial states of the aircraft, we can predict a whole sequence of flight trajectories to the last timestamp by recursively predicting the next state and 4D matching the state with weather information. This task can therefore be viewed as a combination of a translation and a sequence generation problem [18], where an encoder neural network takes an input sequence that corresponds to the last filed flight plan, and a decoder neural network recurrently predicts an output sequence corresponding to an actual flight trajectory.

### 3.3. Data Sources and Preprocessing

In this work, we used four datasets from different sources. The *flight tracks dataset*, which comes from FAA Traffic Flow Management System (TFMS), contains 4D positions of each aircraft throughout its flight – latitude, longitude, altitude, and time, whose resolutions are respectively about 1 minute, 1 minute, 100 feet, and 1 minute. For this study, we limit our scope for flights from George Bush International Airport (IAH) to Logan International Airport (BOS) in 2013, since over 95% of flights are operated by the United Airlines. During the preprocessing, we first excluded flights where spatial or temporal discontinuities were detected, and ones that started or ended outside of the selected terminal areas (0.5-degree latitude/longitude boxes). We then down sampled flight by eliminating one out of every two track points to reduce computational complexity. Lastly, we derived the course, and latitude and longitude speeds by assuming each flight has a constant ground velocity between two consecutive points. The final dataset includes 1,679 flights with an average sequence length of 94.

The *flight plan dataset*, which also comes from FAA TFMS, contains the last filed 2D flight plan coordinates (latitude and longitude) for each flight. While the lengths of flight plans vary from 9 points to 144 points, we can reduce their dimensions significantly by identifying characteristic points. We implemented a variant of the *Approximate Trajectory Partitioning* algorithm proposed by [24], in which we introduced a parameter $\alpha \in [1,2]$ to control the length of the output sequence. The final dataset includes 118 unique flight plans with maximal length of 16.

The *atmospheric datasets*, which contain the wind speed dataset and air temperature dataset, come from the North American Mesoscale (NAM) Forecast system. It produces high-resolution atmospheric information four times a day – 00:00, 06:00, 12:00, and 18:00 UTC – and each prediction cycle provides forecast at hours 0, 1, 2, 3 and the last hour (6) of the cycle. The datasets use Lambert Conformal projection and the original horizontal resolution is 614 by 428, in which the longitude ranges from 152.88W to 49.42W, and latitude ranges from 12.19N to 57.33N. However, since our flight tracks only lie in the continental US, we cropped the original georeferencing system so that the new georeferencing is a box area with four corners: (130W, 22N), (130W, 52N), (64W, 22N), (64W, 52N). The new horizontal resolution is about 413 by 336, which is roughly half of the original. Our final horizontal georeferencing system $G$ is shown by the blue raster in Figure 1, where the red raster is the original georeferencing system. Apart from the horizontal projection, the datasets also include 39 isobaric pressure altitude layers ranging from 50 millibar (about 68,000 ft) to 1,000 millibar (about mean sea level). Therefore, at each forecast time instance, the final datasets have three separate data arrays – westly wind speed, southerly wind speed, and air temperature – each of which has dimension $39 \times 413 \times 336$[1].

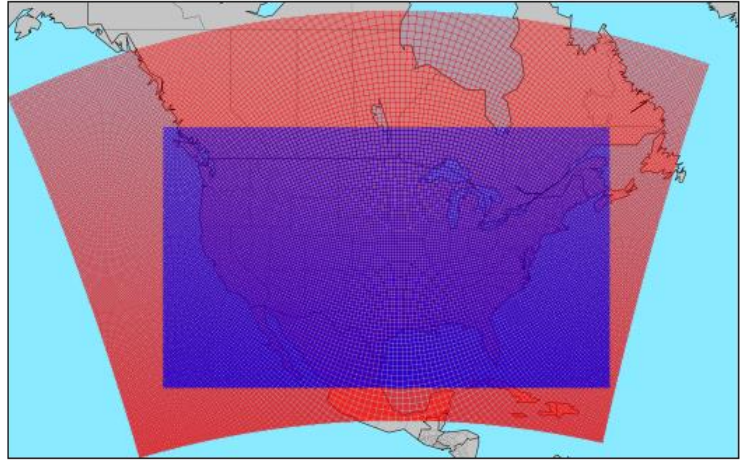

*Figure 1 Georeferencing System G*

The *convective weather dataset* comes from the National Convective Weather Forecast (NCWF) system. In the dataset, every record contains the locations of convective weather polygons (coordinates of boundaries and highest altitude of the storm) and the direction of movement at the time of recording. The dataset was typically updated every 5 minutes. During preprocessing, we first discretized the dataset by the following steps.

(a) Unique storms' altitudes with resolution 1,000 ft, which yield to a list (in 1,000 ft) $L_{alt}^{wx} = [0, 14, 20, 24, 29, 35, 39, 45, 50, 54, 60, 65, 69]$.

(b) Merge storm polygons within the same altitude group and hourly group (i.e., overlay all storm polygons at the same altitude level within an hour).

(c) At each time instance (hour of day), create a binary array with dimension $13 \times 413 \times 336$ [1], where 13 is the number of altitude levels and $413 \times 336$ is the resolution of the horizontal georeferencing system $G$ obtained from processing NAM datasets.

(d) At each time instance, overlay storm polygons at the same altitude level to $G$. If a grid point from $G$ is covered by a storm polygon, then the corresponding element in the data array has value 1, otherwise 0.

---

[1] The new georeferencing system has 138934 horizontal grid points, which is approximately 413 by 336.

## 4. Feature Engineering

In this work, we predict aircraft trajectories using three types of information. The first is the flight's last filed flight plan prior to departure. While this flight plan is clearly an important indicator of what the flight trajectory will be, it is by no means determinative. It is common for aircraft to deviate from their flight plans rather than "fly as filed". Convective weather, winds, clearances to fly direct, and vectors to resolve conflicts can all cause deviations. Figure 2, in which all flights shown by the blue trajectories filed the same flight plan shown by the red curve reveals the extent to which flights deviate from their last pre-departure flight plans.

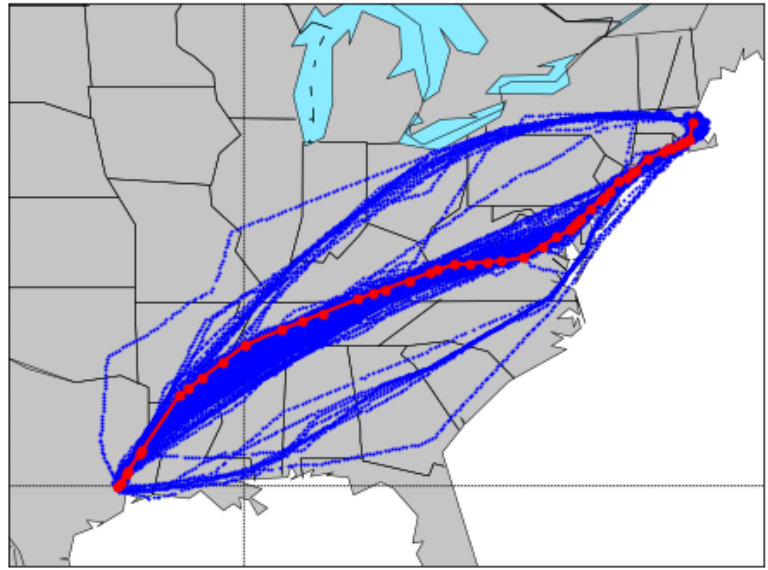

*Figure 2 Actual Flight Trajectories vs Flight Plan(s)*

The second source of information used in our model pertains to convective weather. Strong updrafts and downdrafts are evident within a convection area, which cause significant and unfavorable turbulence. Therefore, aircrafts almost always avoid those areas by either strategically by choosing a route predicted to be clear of convective weather or tactically by maneuvering around convective weather cells. The last category of information pertains to atmospheric conditions, specifically air temperature and wind speeds. The conditions can affect trajectories in at least two ways. First, flights prefer routes with strong tailwind since it saves both fuel and time. Third, high temperatures are associated with more turbulence, which pilots may seek to avoid. Finally, pilots prefer flying through cold and dense air since aircraft engines can produce more power.

To predict actual flight trajectories, it is crucial for us to convert the underlying datasets into features that provide the basis for the subsequent modelling. In this section, we first introduce the concept of *feature cube*, which is a multi-dimensional data array that contains the above weather-related information surrounding a track point. Then we summarize a *batch mode* and a *recursive* *mode* approach to efficiently match the flight track with data cubes during the training and inference process, respectively.

## 4.1. Feature Cube Referencing System

To match flight trajectories with the relevant raw datasets, we generate a 4D referencing system that is related to a given track point. For each track point, we first construct a grid square surrounding the point. The width and height of the grid square are respectively dx and dy degrees in latitude/ longitude, and the resolution of the square is nx by ny points. Due to the fact that only the weather conditions in front of an aircraft matters, we center one side, instead of the centroid, of the grid square at the track point. Then we rotate the grid square by the course of the previous track point. Second, we specify the altitude of the grid square. For the atmospheric datasets (wind speeds and air temperature), we use the closest pressure altitude (e.g., 200 millibar) to the track point's altitude; for the convective weather dataset, we use the closest altitude in $L_{alt}^{wx}$ to the track point's altitude. Lastly, we set the timestamp for the grid square by directly using the corresponding track point's timestamp.

The details of the above process are described in Algorithm 1. Notice that each flight is a sequence of track points, and therefore, has a trajectory of the referencing grids, which we term as *feature cube grid path*. Moreover, since the above operation is independent from track point to track point, we can generate all the grids in a batch mode to speed up the process. An example of the feature cube grid path is illustrated in Figure 3, where the blue series is one of the actual trajectories and the red lattice shows the grid path around the trajectory.

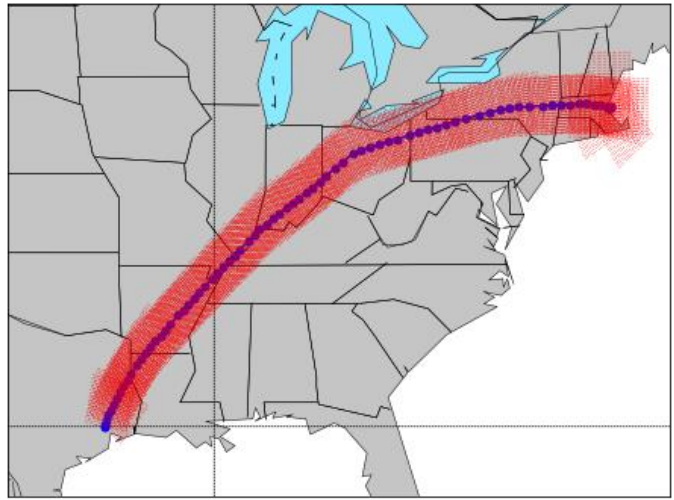

*Figure 3 Feature Cube Grid Path*

| **Algorithm 1. Feature Cube Grid Generator (FCGG)** |
|---|
| INPUTS |
| A preprocessed[1] 4D flight trajectory $TP_j = [p_1, p_2, \ldots, p_T]$; $p_t = [x, y, z, t]$;[2] <br> Atmospheric datasets unique altitude list $L_{alt}^{atm}$; <br> Convective weather dataset unique altitude list $L_{alt}^{wx}$. |
| PARAMETERS |
| Grid longitude size dx = 2°; <br> Grid latitude size dy = 2°; <br> Grid longitude resolution nx = 20; <br> Grid latitude resolution nx = 20. |
| OUTPUTS |
| A set of feature cube grids for all track points on $TP_j$. |
| ALGORITHM |
| Step 1: Create empty feature cube grids. <br>     Create $T$ grid arrays $FG_i, i = \{1,2,\ldots,T\}$, each of which has dimension (nx · ny, 2). <br>     $\delta x = \frac{dx}{nx-1}; \delta y = \frac{dy}{ny-1}$; <br>     $FG_i[0][0] = 0; FG_i[0][1] = -\frac{dy}{2}; FG_i[nx \cdot ny - 1][0] = dx; FG_i[nx \cdot ny - 1][1] = \frac{dy}{2}$; <br>     $FG_i[k][0] = q \cdot \delta x; FG_i[k][1] = -\frac{dy}{2} + p \cdot \delta y$; where $k = p \cdot nx + q, 0 \leq k \leq nx \cdot ny$. <br> Step 2: Create rotation arrays. <br>     Create $T$ rotation arrays $R_i, i = \{1,2,\ldots,T\}$, each of which has dimension (2,2). <br>     $R_i = \begin{bmatrix} \cos(\theta_i) & -\sin(\theta_i) \\ \sin(\theta_i) & \cos(\theta_i) \end{bmatrix}$; where $\theta_i$ is the course for track point $p_i$. <br> Step 3: Rotation. <br>     Rotate 2D feature cube grids and project back to the flight track space. <br>     $FG_i = FG_i \cdot R_i + p_i'$, where $p_i' = [lon, lat]_i$. <br> Step 4: Align altitude and time for flight tracks. <br>     $TP_j^{atm} = [p_1^{atm}, \ldots, p_T^{atm}]$, where $p_t^{atm} = [z^{atm}, t]$, $z^{atm}$ is the closest pressure altitude for $alt_t$ in $L_{alt}^{atm}$. <br>     $TP_j^{wx} = [p_1^{wx}, \ldots, p_T^{wx}]$, where $p_t^{wx} = [z^{wx}, t]$, $z^{wx}$ is the closest pressure altitude for $alt_t$ in $L_{alt}^{wx}$. <br> Step 5: <br>     Return $FG_j = [FG_1, FG_2, \ldots, FG_T]$, $TP_j^{atm}$ and $TP_j^{wx}$. |

*Note:*
1. A preprocessed flight track point contains 4D points, course, latitude speed, and longitude speed.
2. We can run this algorithm in a batch mode by specifying $TP_j$ to be a collection of 4D flight trajectories; we can also run this algorithm in a recursive mode by specifying $TP_j$ to be one single track point.

## 4.2. Feature Cube Matching

To construct highly descriptive feature space, we match the proposed feature cube grid for every track point with the weather-related datasets mentioned above. Both the grids and datasets have 4-dimensional spatiotemporal structures – latitude, longitude, altitude, and time, each grid therefore needs to query into the weather datasets and use the value of the closest 4D reference point as its feature value, which is generally computational extensive. In this study, we proposed a tree-based matching algorithm that efficiently matches flight trajectories with high-dimension datasets in a 4D manner. For conciseness, we only describe an example of matching with convective weather dataset, and the full algorithm is presented in Algorithm 2.

To accomplish our goal, two sets of *k*-d trees ([25]) – spatial tree and temporal tree – are firstly constructed. To be more specific, the spatial tree $TR_S$ is based on the horizontal georeferencing system $G$, where each data entry is a tuple of grid's latitude and longitude. Since both the atmospheric datasets and convective weather dataset share the same $G$, $TR_S$ is static. Also notice that $TR_S$ only has two dimensions and does not include the altitude information. We do this because there is a very limited number of altitude levels from weather datasets, and therefore we can perform quick altitude matching by simply using grouping and indexing. The temporal tree $TR_T$ is a one-dimensional *k*-d tree that constructed based on the elapsed time (e.g., seconds) from a pre-specified baseline time (e.g., 01/01/2013 00:00 Zulu) for each convective weather data entry.

To perform efficient 4D matching, we first generate *feature cube grids* for all flight tracks in a batch mode using Algorithm 1. Second, we batch query into the spatial tree $TR_S$ using the feature cube 2D grids, which returns the indices of the closest grid points from the georeferencing system $G$ to the feature cube grids. Third, we group the feature cube grid by their altitudes. However, since storms usually propagate fast, we should not only consider the convections for the current altitude level, but also convections below and above. Therefore, within each group, we also specify an altitude buffer, and subset all convective weather datasets that are within the altitude range. Fourth, within each group, we further query into the temporal tree $TR_T$ using the feature cube grids timestamps with a maximal time distance bound, and collect the closest time indices from the weather dataset to the feature cube grids. Lastly, we use the three indices – altitude index, 2D spatial index, and temporal index – to collect the matched convective weather, which is an array with dimension $(k, \text{nx}, \text{ny})$, where $k$ is the number of altitude levels within the altitude buffer, for each feature cube grid. Lastly, for each matched feature cube grid, we overlay all $k$ layers of matched weather array so that the resulting array has dimension $(\text{nx}, \text{ny})$, and each element is 1 if there was convection at the corresponding grid and 0 otherwise.

Notice that the above description only presents our matching procedure for convective weather dataset. For atmospheric datasets, however, the process is very similar except that we don't specify altitude buffers. Therefore, the final matched feature cube for each feature cube grid has 4 layers – each with dimension $(\text{nx}, \text{ny})$ – where the first layer is a binary array indicating convective weather, the second layer represents air temperature, the third layer indicates westly wind speed, and the last layer indicates southerly wind speed.

| **Algorithm 2. Feature Cube Matching (FCM)** |
|---|
| INPUTS<br>The 2D georeferencing system $G = [(lon, lat)_1, (lon, lat)_2, \ldots, (lon, lat)_N]$;<br>A feature cube grid path $FG_j, T_j^{atm}, T_j^{wx}$;<br>Atmosperic datasets $D_{atm} = \{D_1^{atm}, D_2^{atm}, \ldots, D_t^{atm}, \ldots, D_M^{atm}\}$, where $D_t^{atm}$ contains wind speeds and air temperature information at time $t$;<br>Convective weather datasets $D_{wx} = \{D_1^{wx}, D_2^{wx}, \ldots, D_t^{wx}, \ldots, D_K^{wx}\}$, where $D_t^{wx}$ contains convective weather information at time $t$;<br>PARAMETERS<br>Baseline time $BT$ = 01/01/2013 00:00 Zulu;<br>Altitude buffer $AB$ = 20000 ft;<br>Maximal time bound $TB$ = 1 hr;<br>OUTPUTS<br>A set of matched feature cubes for the corresponding feature cube grid path. |
| ALGORITHM<br>Step 1: Spatiotemporal Tree Construction.<br>    $TR_S = kdtree(G)$;<br>    $TR_T^{atm} = kdtree(ElapT_{atm})$;<br>    $TR_T^{wx} = kdtree(ElapT_{wx})$;<br>    where $ElapT_{atm} = t_{atm} - BT$; $ElapT_{wx} = t_{wx} - BT$; $t_{atm}$ and $t_{wx}$ are respectively arrays of unique timestamps from $D_{atm}$ and $D_{wx}$.<br>Step 2: Matching with atmospheric datasets.<br>    $Idx_S = TR_S.query(FG_j)$;<br>    $D_{matched}^{atm} = []$;<br>    for $(z^g, tg)$ in $T_j^{atm}$.groupby(altitude): # $tg = [t_1^g, \ldots, t_P^g]$<br>        $Idx_T = TR_T.query(tg, maximal\ bound = TB)$<br>        $D_{matched}^g = D_{atm}[Idx_S, z^g, Idx_T]$ # has shape $[3, nx, ny]$<br>        $D_{matched}^{atm}.append(D_{matched}^g)$<br>Step 3: Matching with convective weather dataset.<br>    $Idx_S = TR_S.query(FG_j)$;<br>    $D_{matched}^{wx} = []$;<br>    for $(z^g, tg)$ in $T_j^{wx}$.groupby(altitude): # $tg = [t_1^g, \ldots, t_P^g]$<br>        $D_{wx}^{sub} = D_{wx}[z^g - AB: z^g + AB]$<br>        $Idx_T = TR_T.query(tg, maximal\ bound = TB)$<br>        $D_{matched}^g = D_{wx}^{sub}[Idx_S, z^g, Idx_T]$ # has shape $[k, nx, ny]$<br>        $D_{matched}^g[m, n] = OR(D_{matched}^g[:, m, n])$ # has shape $[nx, ny]$<br>        $D_{matched}^{wx}.append(D_{matched}^g)$<br>Step 4:<br>    Return $[D_{matched}^{wx}, D_{matched}^{atm}]$ |

## 5. Module Design

Aircrafts that "fly as filed" frequently deviate from their pre-departure last filed flight plans for the various reasons discussed above. To predict such "deviations", or equivalently the actual flight trajectories, we formulate the problem as a combination of translation and sequence generation problem, in which we recurrently use weather information to "translate" a flight plan to an actual flight trajectory. Specifically, our model integrates three modules: (a) an encoder

LSTM to embed flight plans into a fixed-size feature vector; (b) a decoder LSTM that maps the fixed-size feature vector to the target flight trajectory sequence; and (c) a set of convolutional layers that are integrated into the decoder network to condense high-dimension weather-related feature cubes into fixed-size feature representations. Figure 4 shows the general training framework (unrolled LSTM).

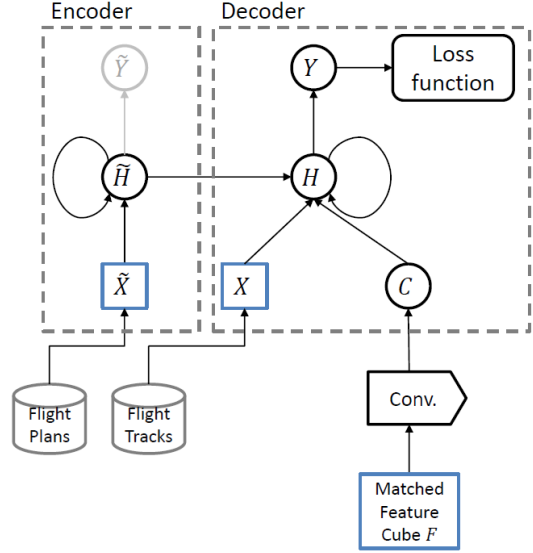

*Figure 4 Training Framework*

## 5.1. Architecture

**Encoder-decoder LSTM**

Long Short-Term Memory (LSTM) networks have been shown to be an effective tool to learn representations from sequential data with long temporal dependencies. Inspired by the success of [18] to solve neural translation problems, we develop an LSTM-based encoder-decoder architecture to map flight plans to actual flight trajectories. The encoder network $\widetilde{LSTM}$ inputs the flight plan sequence $\tilde{X}$ and produces a sequence of hidden states (see Eq. 1). We then obtain the hidden state at the last timestamp $\widetilde{H}_{\tilde{T}}$, which is a fixed-size representation of the flight plan.

$$\widetilde{H}_t = \widetilde{LSTM}(\widetilde{H}_{t-1}, \tilde{X}_t, \widetilde{W}), t = 1, 2, \ldots, \tilde{T} \qquad (1)$$

The decoder network $LSTM$ uses $\widetilde{H}_{\tilde{T}}$ as the initial hidden state and predicts the whole sequence of actual flight states. The actual flight state at time $t$, denoted by $X_t = [x, y, z, \dot{x}, \dot{y}]_t$, is represented by a list of variables that includes latitude, longitude, altitude, latitude speed and longitude speed at time $t$. We assume that the state at each timestamp follows Gaussian mixtures whose parameters are learned by the decoder network. For example, Equation 2 indicates that at timestamp $t$, the decoder network takes the hidden variable at time $t-1$, current aircraft state $X_t$, some weather-related feature representations $C_t$ from the convolutional layers, and the weight variables $W$ to be learned, and outputs the hidden variable $H_t$, which further constitutes the Gaussian mixtures parameters. Eq. 3 suggests that our state variable $X_t$ is drawn from $K$ independent Gaussian distributions, each with a weight $\phi_t^i$. To further simply the model, we assume the covariance matrix for each Gaussian component $\Sigma_t^i$ is block diagonal, with $\Sigma_t^{1,i}$ a full $3 \times 3$ covariance matrix for latitude, longitude, and altitude, and $\Sigma_t^{2,i}$ a full $2 \times 2$ covariance matrix for latitude speed and longitude speed. All the Gaussian mixtures parameters are functions of outputs of the decoder network $H_t$. However, to ensure validity of the distribution (e.g., positive definite covariance matrix), we need to further scale the raw outputs $H_t$. Specifically, the Gaussian component weights $\phi_t^i$ is scaled by a softmax function (Eq. 4), the Gaussian mode vectors $\mu_t^i$ are nonlinear mappings of $H_t$ (Eq. 5, e.g., exponential linear unit activation). To ensure the positive definiteness of the covariance matrix, the decoder network outputs the lower triangular components of matrices $L_t^{1,i}$ and $L_t^{2,i}$ that further construct the covariance matrices by Cholesky decomposition (Eq. 6).

$$H_t = LSTM(H_{t-1}, X_t, C_t, W | \widetilde{H}_{\widetilde{T}}), t = 1, 2, \dots, T$$
$$C_t = \text{convNN}(F_t, W_{cnn}) \tag{2}$$

$$X_t \sim \sum_{i=1}^{K} \phi_t^i \mathcal{N}(\mu_t^i, \Sigma_t^i); \mu_t^i = [\mu_{lat}^i, \mu_{lon}^i, \mu_{alt}^i, \mu_{latspd}^i, \mu_{lonspd}^i]_t;$$
$$\Sigma_t^i = \begin{bmatrix} \Sigma_t^{1,i} & 0 \\ 0 & \Sigma_t^{2,i} \end{bmatrix}; \Sigma_t^{1,i} = \Sigma_{(lat,lon,alt),t}^i; \Sigma_t^{2,i} = \Sigma_{(latspd,lonspd),t}^i \tag{3}$$

$$[\phi_t^1, \phi_t^2, \dots \phi_t^K] = \text{softmax}(W_\phi \cdot H_t) \tag{4}$$

$$[\mu_t^1, \mu_t^2, \dots \mu_t^K] = f(W_\mu \cdot H_t) \tag{5}$$

$$\Sigma_t^{1,i} = L_t^{1,i} \cdot L_t^{1,i^T}; \Sigma_t^{2,i} = L_t^{2,i} \cdot L_t^{2,i^T};$$
$$[L_t^{1,i}, L_t^{2,i}]_{i=1,2,\dots,K} = g(W_L \cdot H_t) \tag{6}$$

The weights of our neural networks are learned by optimizing the loss function of our model, which is the negative log likelihood and can be formulated by Eq. 7. In the equation, the probability function $P(\cdot)$ is the density function of a normal distribution.

$$L(\widetilde{W}, W, W_{cnn}, W_\phi, W_\mu, W_L) = -\sum_{t=1}^{T} \log \left[ \sum_{i=1}^{K} \phi_t^i \cdot P(X_t | \mu_t^i, \Sigma_t^i) \right] \tag{7}$$

**Convolutional layers**

Convolutional neural network (CNN) has achieved great success in image recognition ([26], [27], [28], [29]) and object detection ([30], [31], [32]). As is also well known, deep CNN is not only capable of classification tasks, but also learning feature representations from high-dimensional input signals. Using the feature engineering techniques introduced in Section 4, each actual track point $p_t$ will be matched with a high-dimension feature cube $F_t$. The feature cube can be treated as a multi-channel "image" whose width and height are decided by the size of the relevant regions around the aircraft location, and number of channels is the number of weather-related features considered. In this work, the feature cube has four channels – westerly and southerly wind speeds, air temperature, and convective weather. To extract feature representation from the feature cube, we employ multiple convolutional layers. Specifically, at each timestamp, our convolutional layers use small filters to abstract a feature cube into a fixed-size feature vector and directly feed to the decoder LSTM network ($C_t = \text{convNN}(F_t, W_{cnn})$). During the training process, therefore, the loss will be back propagated to all weights of the convolutional layers.

## 5.2. Inference Process

During the inference time, we are trying to answer to what the rest of trajectory will be, if we know an aircraft's last filed flight plan, first $T'$ states ($1 \leq T' \leq T$), and the corresponding $T'$ weather-related feature cubes. One simple solution is to directly feed the flight plan $\tilde{X}_t$, $t = 1, 2, \ldots, \tilde{T}$, first $T'$ states and feature cubes $X_t, F_t, t = 1, 2, \ldots, T'$ to the trained networks, and obtain the set of Gaussian mixtures parameters $[\phi_t^i, \mu_t^i, \Sigma_t^i]$, $i = 1, 2, \ldots, K$. Then we use the Gaussian mixtures to sample one aircraft state, say $\hat{X}_{T'+1}$, and use Algorithms 1 and 2 (in *recursive mode*) to match the generated point with weather data and obtain $F_{T'+1}$. By recursively repeating the process until the last timestamp $T$, we can generate a whole flight trajectory after time $T'$. Although simple as it is, the predicted sequence has very large variance, especially towards the end of the sequence. One may argue that this can be tackled by sampling multiple points, however, in practice, this ends up with "zigzagged" predictions due to the symmetry of Gaussian distributions.

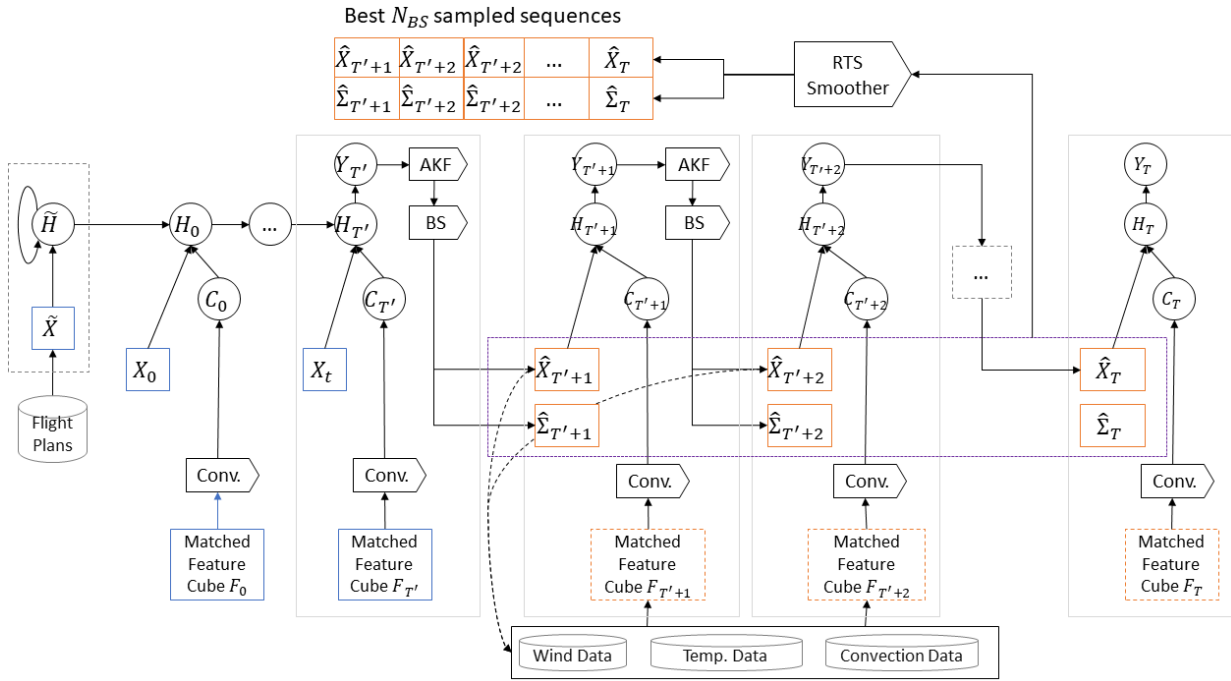

*Figure 5 Inference Framework*

| **Algorithm 3. Adaptive Kalman Filter (AKF)** |
|---|
| INPUTS |
| Predictions (mean and covariance) at timestamp $t-1$: $\hat{X}_{t-1}, \hat{\Sigma}_{t-1}$; |
| Measurements (mean and covariance) of $t$: $X_t, \Sigma_t$; |
| PARAMETERS |
| Aircraft dynamic matrix $A$; |
| Initial process noise covariance matrix $Q$; |
| Outlier gating $e_1$; Maneuver gating: $e_2$; |
| Maneuver scaling factor: $Q_s$. |
| OUTPUTS |
| Kalman filter estimates at timestamp $t$: $\hat{X}_t, \hat{\Sigma}_t$. |
| ALGORITHM |
| Step 1: Kalman update<br>    Predict<br>$$\underline{X}_t = A \cdot \hat{X}_{t-1}$$ $$\underline{\Sigma}_t = A \cdot \hat{\Sigma}_{t-1} \cdot A^T + Q_t; \; Q_t = Q$$<br>    Update<br>$$S_t = H \cdot \underline{\Sigma}_t \cdot H^T + \Sigma_t$$ $$K_t = \underline{\Sigma}_t \cdot H^T \cdot S_t^{-1}$$ $$R_t = H \cdot (X_t - \underline{X}_t)$$ $$\hat{X}_t = \underline{X}_t + K_t \cdot R_t$$ $$\hat{\Sigma}_t = (I - K_t \cdot H) \cdot \underline{\Sigma}_t$$<br>Step 2: Gating<br>    If $|R_t| > e_1$, $M = 1$, return $\underline{X}_t, \underline{\Sigma}_t, M$.<br>    Else if $|R_t| > e_2$, set $Q_t = Q_s \cdot Q$, and go to step 1 and collect $\hat{X}_t, \hat{\Sigma}_t$, $M = 0$, return $\hat{X}_t, \hat{\Sigma}_t, M$.<br>    Else, $M = 0$, return $\hat{X}_t, \hat{\Sigma}_t, M$. |

Therefore, instead of using sampling techniques in most literatures ([22], [23]), we directly use the mean vector ($\mu_t^i$) of each Gaussian component, and the output sequence is the one with the best cumulative log likelihood value among all generated sequences. Figure 5 shows the overview of the proposed inference procedure. To be more specific, at timestamp $t, t > T'$, we first perform a feed forward and collect parameters of the Gaussian mixtures $[\phi_t^i, \mu_t^i, \Sigma_t^i]$, $i = 1,2,...,K$. Second, we use Eq. 8 to calculate a cumulative log likelihood for each Gaussian component, where function $l(\cdot)$ gives a negative value if $\mu_t^i$ is considered as an outlier by the subsequent steps (adaptive Kalman Filter) since sequence with outlier predictions is less favorable. We assign weights ($\pi_1$ and $\pi_2$) to different terms in the log likelihood calculation, this is because we find it more important to choose the correct Gaussian component in practice. We also point out that given one aircraft state and feature cube at a timestamp, the trained neural network will generate $K$ estimates from Gaussian mixtures. Therefore, for each newly generated state estimate, we need to trace back its parent state and cumulate the log likelihood for the whole trajectory ($L_{t-1}^J$). Third, if we denote the $i^{th}$ predicted state and covariance at timestamp $t-1$ as $(\hat{X}_{t-1}^i, \hat{\Sigma}_{t-1}^i)$, then for each predicted $(\mu_t^i, \Sigma_t^i)$ tuple at timestamp $t$, we use an adaptive Kalman Filter (AKF) with gating in order to further improve the predictive power (Algorithm 3). Eq. 9 illustrates this step, where $M_t^i$ is an indicator variable signifying whether the input mean vector $\mu_t^i$ is an outlier or not. At each timestamp we generate $K$ new state estimates, the total number of generated trajectories grows exponentially ($K^T$). Accordingly, in the fourth step, we rank $L_t^i$ in a decreasing order and only keep the largest $N_{BS}$ sequences (a.k.a., beam search). Then we use Algorithm 1 and 2 (in *batch mode*) to match the best $N_{BS}$ aircraft state estimates $\hat{X}_t^i$ with weather features and perform feed forward to generate $[\phi_{t+1}^i, \mu_{t+1}^i, \Sigma_{t+1}^i]$. We repeat the above steps recursively until the last timestamp $T$. Lastly, we pick the sequence out of $N_{BS}$ sequences with the best $L_t^i$ and use Rauch-Tung-Striebel smoother ([33]) to further improve the prediction. The full inference algorithm is described Algorithm 4.

$$L_t^i = L_{t-1}^J + \pi_1 \log \phi_t^i + \pi_2 \log P(\mu_t^i | \mu_t^i, \Sigma_t^i) + l(M_t^i), \pi_1 + \pi_2 = 1 \tag{8}$$

$$(\hat{X}_t^i, \hat{\Sigma}_t^i, M_t^i) = \text{AKF}(\hat{X}_t^i, \hat{\Sigma}_t^i, \mu_t^i, \Sigma_t^i) \tag{9}$$

| **Algorithm 4. Trajectory Generator (TG)** |
|---|
| INPUTS |
| Trained neural networks $NN$. |
| Last filed flight plan sequence $\tilde{X} = [\tilde{X}_t], t = 1,2, \ldots, \tilde{T}$; |
| Initial actual flight state trajectory $X = [X_t], t = 1,2, \ldots, T'$; |
| Matched feature cube path $F = [F_t], t = 1,2, \ldots, T'$; |
| Weather-related datasets $D_{atm}$ and $D_{wx}$. |
| PARAMETERS |
| Log likelihood weights $\pi_1, \pi_2$; |
| Beam search size $N_{BS}$; |
| Length of actual trajectories $T$; |
| Outlier loss function $l(\cdot)$. |
| OUTPUTS |
| Sampled best $N_{BS}$ actual flight trajectories. |
| ALGORITHM |
| Step 1: Feed forward. |
|     Feed forward $\tilde{X}, X, F$ to the trained neural networks and collect the predicted Gaussian mixtures parameters at the last timestamp $[\phi_t^i, \mu_t^i, \Sigma_t^i], i = 1,2, \ldots, K, t = T'$. |
| Step 2: Filtering |
|     For the $i^{th}$ Gaussian component |
|         $t = t + 1; X_t = \mu_{t-1}^i; \Sigma_t = \Sigma_{t-1}^i$; |
|         $\hat{X}_t, \hat{\Sigma}_t, M_t = AKF(X_t, \Sigma_t, \hat{X}_{t-1}, \hat{\Sigma}_{t-1})$; |
|         $L_t^i = \pi_1 \log \phi_t^i + \pi_2 \log P(X_t | X_t, \Sigma_t) + l(M_t)$; |
|         Trace parent trajectory with cumulative log likehood $L_{t-1}^J$, and assign $L_t^i = L_t^i + L_{t-1}^J$; |
|     Rank $L_t^i$ in a decreasing order and pick the first $N_{BS}$ state estimates $\hat{X}_t = [\hat{X}_t^j], j = 1,2, \ldots, N_{BS}$. |
| Step 3: Matching |
|     $F_t = FCM(\hat{X}_t, D_{atm}, D_{wx})$; |
| Step 4: Recursion |
|     Repeat step 1 to 3 until reaching the end timestamp $T$. |
|     Return the best $N_{BS}$ flight trajectories. |

## 5.3. Implementation Details

**Feature Engineering**

In our feature engineering process, the horizontal size of each feature cube grid is chosen as $2° \times 2°$ latitude-longitude with resolution $20 \times 20$, which approximately covers a 120-nautical-mile by 120 nautical-mile square region in front of the aircraft's location. As a result, the dimension of each feature cube is $20 \times 20 \times 4$, where the last dimension is constituted by westly and southerly wind speeds, air temperature, and convective weather. All the features except the convective weather are normalized to zero mean and unit variance before feeding into the neural networks.

To potentially transfer our model to a variety of airport pairs, we also subtract the coordinates (latitude, longitude, altitude) of the origin airport from both the last filed flight plans and actual flight tracks and use the outcomes in both the training and inference process. Similar to the feature engineering, we normalize those outcomes to zero mean and unit variance.

**Training**

For the encoder network, we first use an embedding layer with dimension 32 for the flight plan coordinates before feeding them to the LSTM. Our encoder LSTM has two layers, and each layer has a fixed-size hidden state with dimension 128. $K = 3$ Gaussian mixture components are chosen to model the probabilistic distributions of actual flight tracks. Therefore, at each timestamp, the number of decoder LSTM output parameters is 45.

The convolutional layers in the decoder network learn representations from the weather feature cubes. Four layers – three convolutional and one fully-connected – are proposed and the overall structure is summarized in Figure 6. The first layer inputs the feature cube with 16 filters of size $6 \times 6 \times 4$ and stride 2. The second layer uses 16 filters of a smaller size $3 \times 3 \times 16$ and stride 1, and the third layer uses 32 filters of $3 \times 3 \times 16$ and stride 1. The fully connected layer has 32 neurons. Notice that pooling and padding operations are not employed since locational information of weather is important in our problem domain.

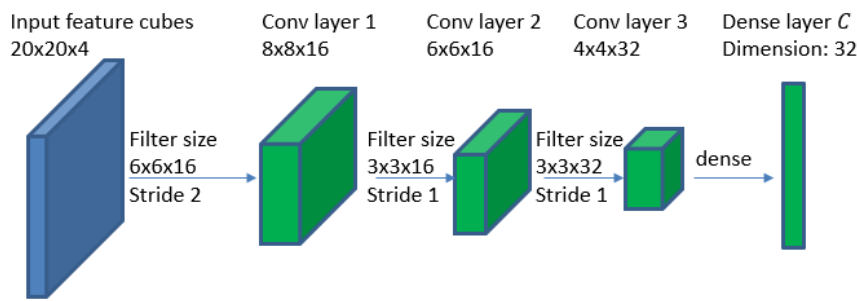

*Figure 6 Structure of the Convolutional Layers*

The decoder network is also a two-layer LSTM with 128-dimensional hidden state and takes the last hidden state from the encoder LSTM as its initial state. Feature representations from the convolutional layers and aircraft state variables are fed into an embedding layer with 64 neurons before entering the decoder LSTM. The outputs of the LSTM will then be mapped to the estimates of parameters of Gaussian mixtures (3 components) using a dense layer. In our architecture, we choose the exponential linear unit (ELU) function for all activations.

Lastly, we use the Nesterov Momentum optimizer [34] with gradient clipping and batch size 256 to train our neural networks. The learning rate is initialized as 0.001 and is decaying every 1,000 epochs.

**Inference**

In calculating the cumulative log likelihood, we assign a higher weight to the probability ($\phi_t^i$) of Gaussian component, specifically, $\pi_1 = 0.8, \pi_2 = 0.2$.

In the adaptive Kalman filter, we assume a simple linear dynamic (Eq. 10) where at each timestamp the aircraft moves at a constant horizontal speed and zero vertical speed. The changes in horizontal speed and altitude are captured by the measurement (predictions from trained neural networks) and are treated as the errors in the linear dynamic system. The process error $Q$ is assumed to be a diagonal matrix in Eq. 11. We choose two gating thresholds $e_1 = 0.8, e_2 = 0.3$ to signify whether the measurement is an outlier, or a maneuver, or neither. Namely, if the sum of absolute errors of latitude and longitude is larger than $e_1$, then the measurement is an outlier and a negative value $l(M = 1) = -9$ will be added to the trajectory's cumulative log likelihood. If the error is greater than $e_2$, then the measurement is considered as a maneuver and we increase the process error $Q$ by a factor of $Q_s = 10$ and repeat the Kalman filter process. Lastly, to ensure the numerical stability, we only use the diagonal of the matrix $S_t$ in calculating the inverse of the updated covariance matrix.

$$\underline{X_t} = A \cdot \hat{X}_{t-1}; \ A = \begin{bmatrix} 1 & 0 & 0 & \Delta t & 0 \\ 0 & 1 & 0 & 0 & \Delta t \\ 0 & 0 & 1 & 0 & 0 \\ 0 & 0 & 0 & 1 & 0 \\ 0 & 0 & 0 & 0 & 1 \end{bmatrix}; \Delta t = 120 \ sec \tag{10}$$

$$Q = diag\{10^{-3}, 10^{-3}, 1, 10^{-6}, 10^{-6}\} \tag{11}$$

In the beam search, we pick the size $N_{BS} = K^2$ during the sampling procedure, and only output one trajectory with the highest log likelihood.

## 6. Experimental Results

We applied our method on a historical flight trajectory dataset from IAH to BOS in the year 2013. The preprocessed dataset contains 1679 flights and is split into two sets, with 80% in the training set and the rest in the evaluation set. In the inference process, we use the first 20 actual track points and their corresponding feature cubes for every flight on the evaluation set as the observed sequence and predict the rest of flight tracks using Algorithm 4. Figure 7 illustrates two examples of our sampled trajectories, where the red curve is the last filed flight plan, the green curve represents the first 20 observed flight tracks, the magenta curve is the predicted flight tracks, and the blue dashed curve is the actual flight tracks (ground truth). In the figure, the background color indicates the average air temperature, with warmer color (towards red) as higher temperature. The arrows represent wind directions and speeds, and the red polygons are the convective weather regions. The green band indicates the path of 99.7% confidence intervals (three standard errors) around predicted flight tracks. In the figure, the left subplot shows the case with slight convective activities, while the right subplot with substantial convections. Both predicted trajectories agree largely with the actual flown tracks (ground truth), with small deviations in the middle parts of the trajectories. However, those deviations are covered by the prediction intervals, which are those narrow green bands in the plots.

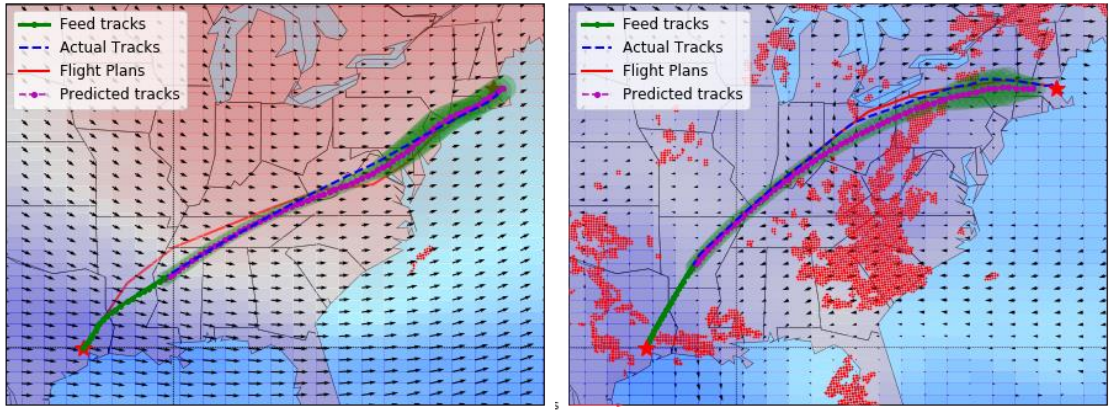

*Figure 7 Examples of our Predictions*

We report four errors below to evaluate the performance of the proposed model. Notice that the point-wise vertical error is signed error while the other three errors are unsigned.

- Point-wise horizontal error (PHE). The distance (in nautical mile) between every predicted point's 2D coordinate (latitude and longitude) and the ground truth.
- Point-wise vertical error (PVE). The vertical distance (in feet) between every predicted point's altitude and the ground truth.
- Trajectory-wise horizontal error (THE). The average point-wise horizontal error (in nautical mile) along each trajectory.
- Trajectory-wise vertical error (TVE). The average point-wise vertical error (in feet) along each trajectory.

Figure 8 shows the histograms of the above errors, and Table 1 presents the average absolute values of the above four errors, denoted respectively as MAPHE, MAPVE, MATHE, MATVE. The distribution of the point-wise horizontal errors is largely skewed towards the left, with an average of 49.60 nautical miles. The point-wise vertical errors, which are signed value, are mostly centered in the range from -5000 ft to 5000 ft. The trajectory-wise horizontal and vertical errors, while both skewed to the left, have similar average absolute values with point-wise errors. Lastly, we point out that we do observe large prediction errors for flights that are have very unusual departure procedures (outlier flights), which requires further research to explore possible solutions.

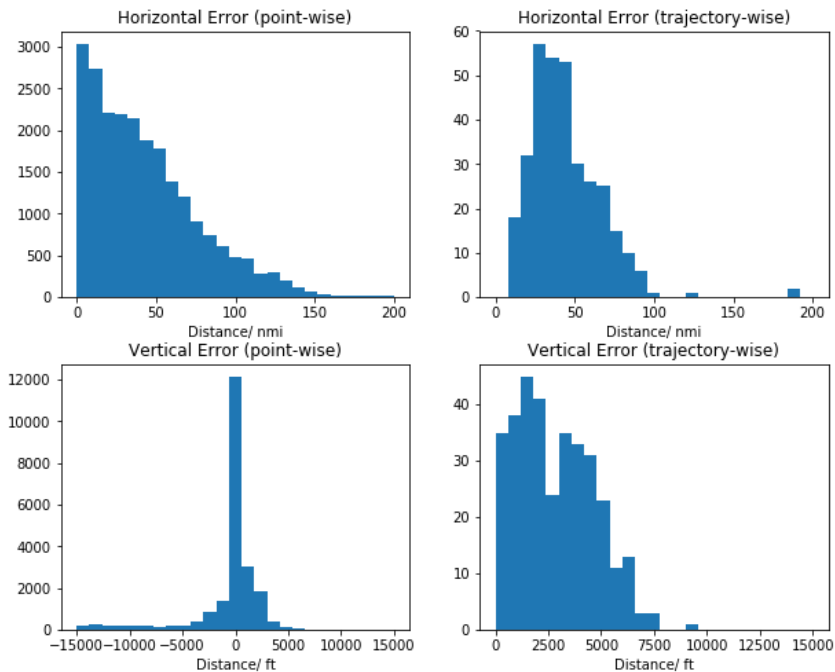

*Figure 8 Prediction Errors*

*Table 1 Average of the Four Absolute Errors*

|  | MAPHE (nmi.) | MAPVE (ft.) | MATHE (nmi.) | MATVE (ft.) |
|---|---|---|---|---|
| Value | 49.60 | 2835.07 | 49.65 | 2861.38 |

We end this section by visualizing the convolutional layers of our neural networks. Figure 9 shows 32 random input feature cubes, with each column a unique feature cube. The first row of the figure represents the convective weather layer, in which the red polygons identify the location of convections; the second row shows the air temperature layer; and the last two rows show respectively the westly and southerly wind speed layer. In the last three layers, warmer color (towards red) indicates larger numerical value. We feed those random inputs into our convolutional layers of the neural networks and obtain the outputs. Figure 10 and Figure 11 show respectively the outputs from the first and second convolutional layers, in which each column represents the corresponding output from Figure 9, and each row shows the results from different filters in the convolutional layers. We first notice that conv layer 1 identifies the locations of convective weather, especially filter 1, 2, 7 and 8. Filters 11, 12 and 16 reflect the two wind speed components, and other filters capture the nonlinear relations among the four input feature cube channels. Figure 11 shows the outputs from conv layer 2, which further abstracts the feature space by introducing more nonlinearities.

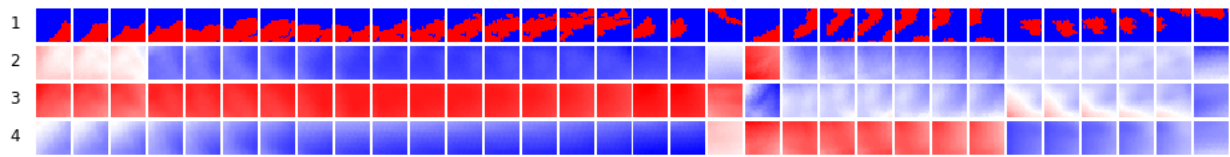

*Figure 9 Input Feature Cubes*

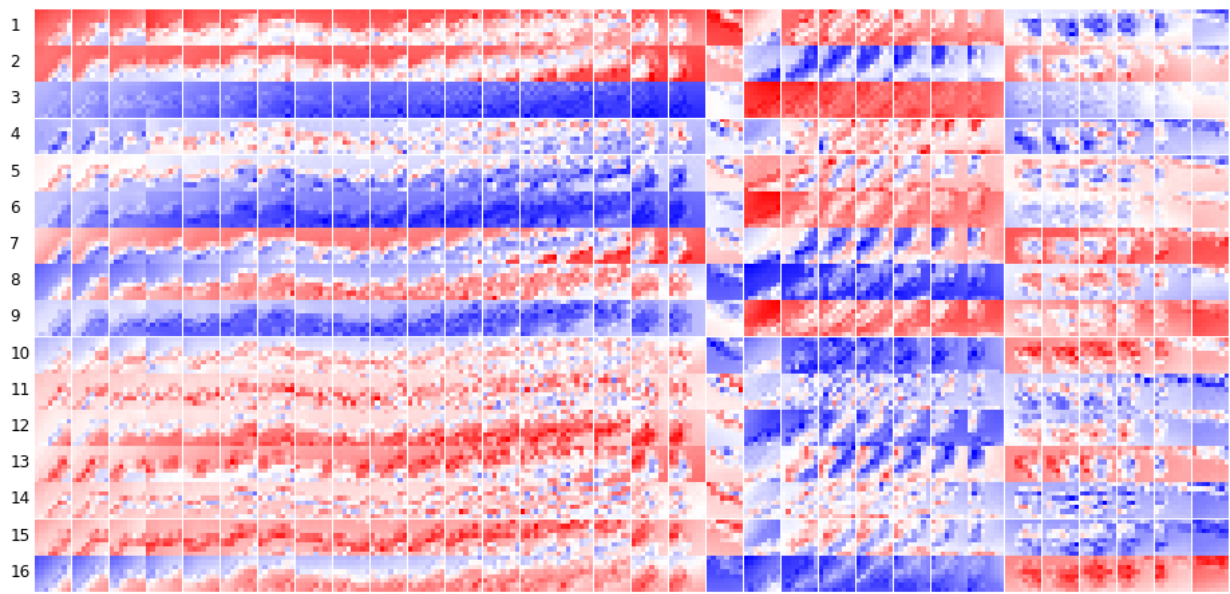

*Figure 10 Outputs from the First Convolutional Layer*

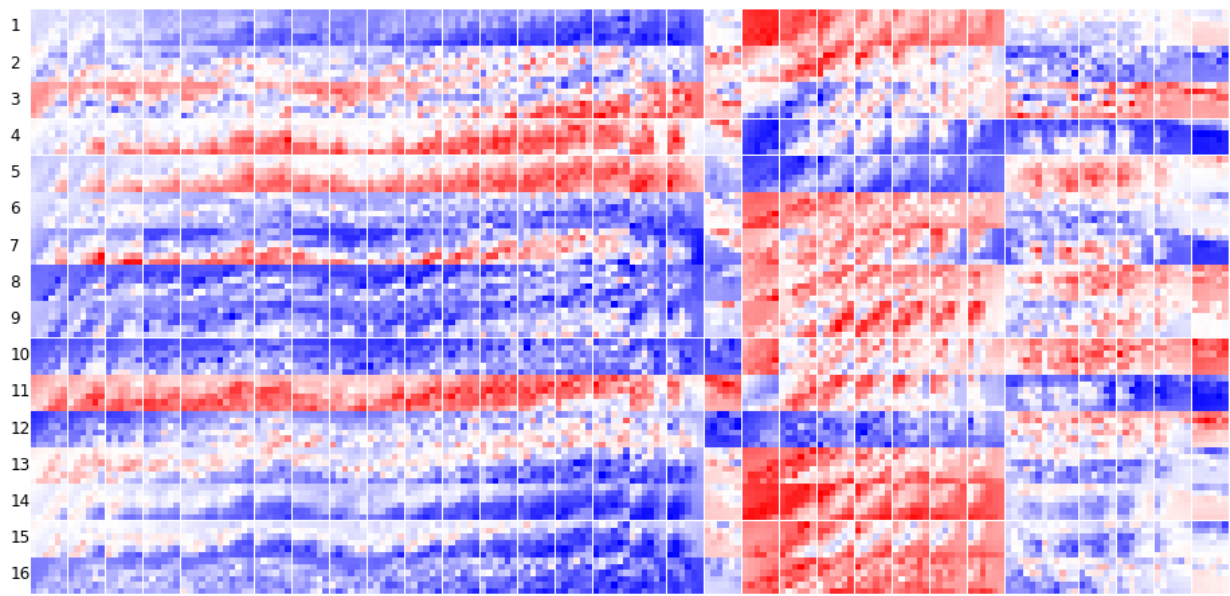

*Figure 11 Outputs from the Second Convolutional Layer*

# 7. Conclusions

In this research, we propose a novel approach to predict the actual aircraft 4D trajectories, using high-dimension meteorological features and last filed flight plans. To tackle this "sequence to sequence" problem, our approach is composed of a matching algorithm, a deep generative model, a training framework, and an inference framework.

The matching algorithm, which is based on the $k$-d tree, efficiently 4D matches actual flight tracks with high-fidelity weather-related datasets – convective weather, air temperature, and wind speeds – in the vicinity of the aircraft. The constructed feature space for each track point is based on the concept of *feature cube grid*, which is a spatiotemporal 4D grid surrounding the track point. By querying such grid into the weather datasets, our matching algorithm constructs "image-like" feature cubes, whose width and height are the size of the interested regions around aircraft locations, and number of channels are the number of weather-related features.

In our approach, we model the actual flight states – latitude, longitude, altitude, latitude speed, and longitude speed – as conditional Gaussian mixtures with parameters to be learned from our proposed deep generative model, which consists of a multi-layer encoder LSTM, a multi-layer decoder LSTM, and a set of convolutional layers. The encoder network inputs the last filed flight plan sequence and produces a fixed-size hidden state variable that is later fed into the decoder network. The convolutional layers learn the feature representations from the matched feature cubes. The convolutional layers are integrated into the decoder network so that the loss will be back propagated to their weights. The decoder network takes the encoder's output as its initial hidden state and predicts the Gaussian mixtures parameters based on each timestamp's flight state and feature representation. In the training process, the loss function is the negative log likelihood.

In the inference process, we first feed the flight plan, observed flight states and their corresponding feature cubes into the trained model and obtain the Gaussian mixtures parameters for the first predicted flight state. Thereafter, we recursively apply an adaptive Kalman filter with gating, which improves the prediction power, and our matching algorithm, which produces the feature cubes for the predicted states, and feed the outputs to the decoder LSTM to predict the rest of the flight trajectory. A beam search algorithm is implemented to reduce computational complexity in the recursion, and a RTS smoother is used to further improve the best predicted trajectory.

We apply our model on the datasets for flights from IAH to BOS in the year 2013. 1342 flights are used for training and 337 flights are used for evaluation. By visualizing our convolutional layers, we observe that the learned filters successfully locate the convective weather and generalize the weather-related features well. We use four metrics to measure the prediction error. Both the point-wise and trajectory-wise average absolute horizontal errors are around 50 nautical miles, while the average absolute vertical errors are around 2800 feet. We also observe large prediction errors for (outlier) flights with unusual departure procedures, which will be explored in our future research. Other future works include extending our matching algorithm to more features such as air traffic management initiatives (miles-in-trail, airspace flow program, etc.) and neighboring aircrafts.